# MedBench: A Comprehensive, Standardized, and Reliable Benchmarking System for Evaluating Chinese Medical Large Language Models

Mianxin Liu†, Jinru Ding†, Jie Xu†, Weiguo Hu†, Xiaoyang Li, Lifeng Zhu, Zhian Bai, Xiaoming Shi, Benyou Wang, Haitao Song, Pengfei Liu, Xiaofan Zhang, Shanshan Wang, Kang Li, Haofen Wang, Tong Ruan, Xuanjing Huang, Xin Sun, Shaoting Zhang*

**Abstract:** Ensuring the general efficacy and goodness for human beings from medical large language models (LLM) before real-world deployment is crucial. However, a widely accepted and accessible evaluation process for medical LLM, especially in the Chinese context, remains to be established. In this work, we introduce “MedBench", a comprehensive, standardized, and reliable benchmarking system for Chinese medical LLM. First, MedBench assembles the currently largest evaluation dataset (300,901 questions) to cover 43 clinical specialties and performs multi-facet evaluation on medical LLM. Second, MedBench provides a standardized and fully automatic cloud-based evaluation infrastructure, with physical separations for question and ground truth. Third, MedBench implements dynamic evaluation mechanisms to prevent shortcut learning and answer remembering. Applying MedBench to popular general and medical LLMs, we observe unbiased, reproducible evaluation results largely aligning with medical professionals' perspectives. This study establishes a significant foundation for preparing the practical applications of Chinese medical LLMs. MedBench is publicly accessible at https://medbench.opencompass.org.cn.

**Keywords:** Medical Large Language Model, Benchmark, Platform, Chinese, Open-source

- Mianxin Liu, Jinru Ding, Jie Xu, Xiaoming Shi, and Shaoting Zhang are with Shanghai Artificial Intelligence Laboratory, Shanghai 200232, China. E-mail: liumianxin@pjlab.org.cn, dingjinru@pjlab.org.cn, xujie@pjlab.org.cn, xiaoming134326@gmail.com, zhangshaoting@pjlab.org.cn.
- Weiguo Hu, Xiaoyang Li, Lifeng Zhu, and Zhian Bai are with Ruijin Hospital Affiliated to Shanghai Jiaotong University School of Medicine, Shanghai 200025, China. E-mail: wghu@rjh.com.cn, lxy11811@rjh.com.cn, zlf@rjh.com.cn, bza@rjh.com.cn.
- Benyou Wang is with Shenzhen Research Institute of Big Data and also with the Chinese University of Hong Kong, Shenzhen 518172, China. Email: wangbenyou@cuhk.edu.cn.
- Haitao Song is with the Shanghai Artificial Intelligence Research Institute and also with Shanghai Jiao Tong University, Shanghai 200240, China. Email: allen5@sjtu.edu.cn.
- Pengfei Liu is with the School of Electronic Information and Electrical Engineering, Shanghai Jiao Tong University, Shanghai 200240, China. E-mail: pengfei@sjtu.edu.cn.
- Xiaofan Zhang is with Qing Yuan Research Institute, Shanghai Jiao Tong University, Shanghai 200240, China. Email: xiaofan.zhang@sjtu.edu.cn.
- Shanshan Wang is with Shenzhen Institutes of Advanced Technology, Chinese Academy of Sciences, Shenzhen 518055, China. Email: ss.wang@siat.ac.cn.
- Kang Li is with Sichuan University West China Hospital, Chengdu 610041, China. E-mail: likang@wchscu.cn.

● Haofen Wang is with the College of Design and Innovation, Tongji University, Shanghai 200092, China. E-mail: haofen.wang@tongji.edu.cn.
● Tong Ruan is with the Department of Computer Science and Technology, East China University of Science and Technology, Shanghai 200237, China. E-mail: ruantong@ecust.edu.cn.
● Xuanjing Huang is with the School of Computer Science, Fudan University, Shanghai 200433, China. E-mail: xjhuang@fudan.edu.cn.
● Xin Sun is with Xinhua Hospital Affiliated to Shanghai Jiaotong University School of Medicine, Shanghai 200092, China. Email: sunxin@xinhuamed.com.cn.
† Mianxin Liu, Jinru Ding, Jie Xu, and Weiguo Hu contributed equally.
* To whom correspondence should be addressed.

## 1 Introduction

Large Language Models (LLMs) play an increasingly critical role across various fields and can potentially reform the healthcare sector. Medical Large Language Models (MLLMs) have thus emerged as a key area of focus [1, 2]. To ensure the reliability of MLLMs before they are deployed in real-world settings, a thorough evaluation is essential [3]. It is vital to establish a comprehensive, standardized, and reliable benchmarking system to assess the general efficacy of MLLMs. Despite this need, there is a notable absence of a universally recognized benchmarking framework, especially in the context of the Chinese context.

There has been a great endeavor to establish Chinese benchmarks for MLLM, such as National Medical Licensing Examination in China-Question Answering (MLEC-QA) [4], Chinese National Medical Licensing Examination (CMExam) [5], Chinese Biomedical Language Understanding Evaluation (CBLUE) [6], and Comprehensive Medical Benchmark in Chinese (CMB) [7]. However, existing benchmarks exhibit several limitations that hinder their suitability for evaluating MLLMs comprehensively. ***Firstly***, existing benchmarks mainly focus on general clinical knowledge and often fall short in providing sufficient coverage across medical specialties, limiting their applicability to diverse healthcare sectors, as evidenced by the reported subpar performance of LLMs in specific fields like nephrology [8] and myopia care [9]. It highlights the necessity of benchmarks that encompass a broader spectrum of medical specialties. ***Secondly***, while existing benchmarks contribute valuable datasets, they neglect the establishment of a standardized evaluation infrastructure. Currently, the benchmark procedure is often chosen and operated by the user, resulting in inconsistent evaluation results for the same dataset. This inconsistency underscores the necessity for not only comprehensive datasets but also a standardized, publicly accessible evaluation infrastructure. ***Thirdly***, the reliability of existing benchmarks poses challenges – The datasets, being entirely open and static, are susceptible to issues like shortcut learning [10] and answer leakage [11], which potentially inflates results artificially. In principle, a dynamic evaluation, with physical separation of question and ground truth, is expected to address this issue.

In response to these challenges, this work introduces MedBench, a pioneering benchmarking system tailored specifically for evaluating Chinese MLLMs. MedBench overcomes the limitations of previous benchmarks by providing comprehensive evaluations across 43 clinical specialties. The evaluated dimensions involve medical language understanding, generation, knowledge question answering, complex reasoning, and healthcare safety and ethics. MedBench employs a fully automated, cloud-based infrastructure to address the standardization and answer leakage issues. It enhances reliability through dynamic evaluation mechanisms, such as circular shuffling on choices and random prompt matching. The subsequent sections elaborate on the development and features of MedBench, highlighting its significance in advancing the evaluation of Chinese MLLMs and bridging current gaps in benchmarking methodologies.

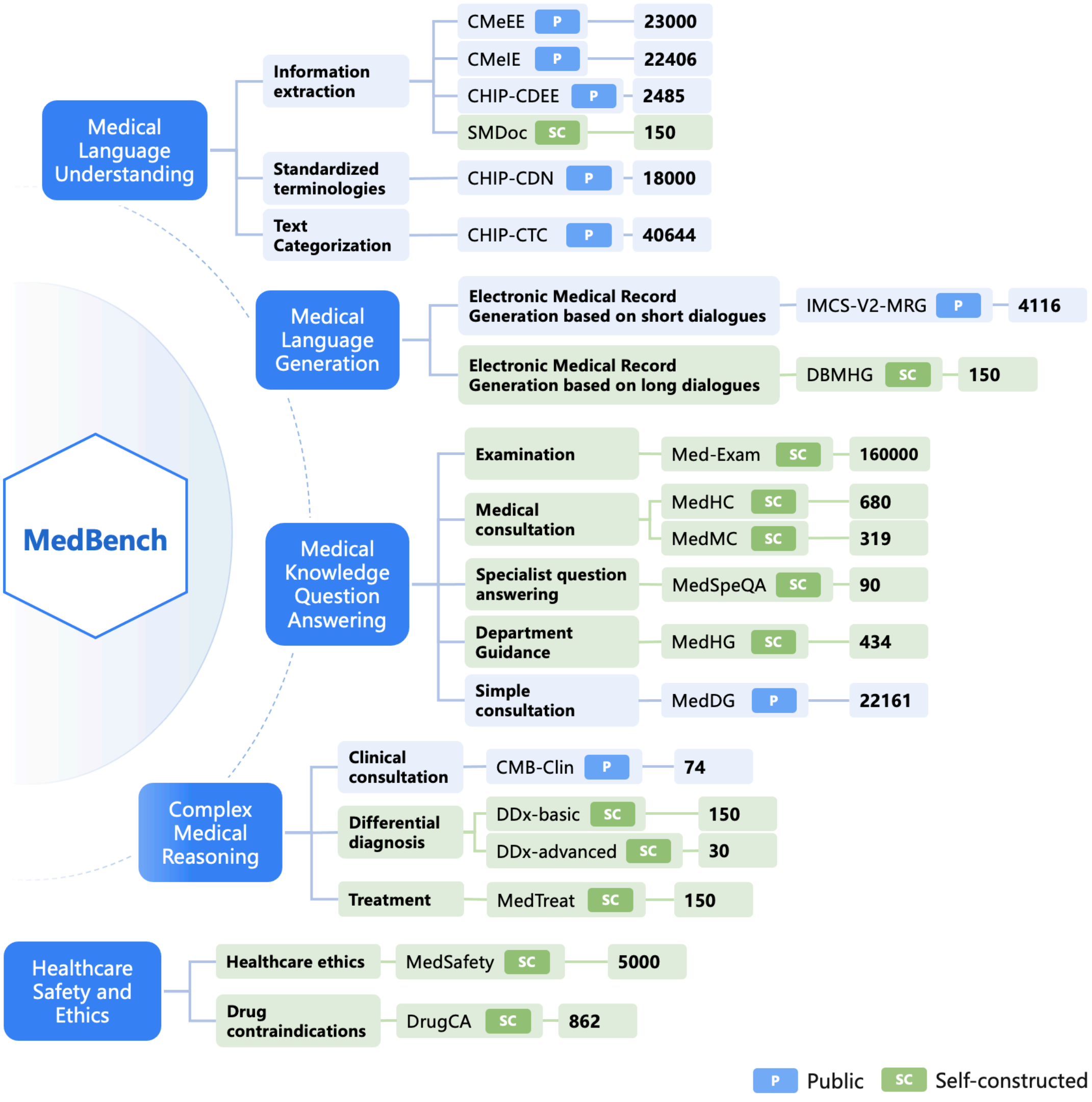

**Figure 1**. The distributions and the measuring dimensions of the datasets in MedBench.

## 2 Method and Material

### 2.1 Benchmark dataset construction

In order to comprehensively assess the capabilities and performance of LLMs in medical field, we aim to measure five evaluation dimensions in MedBench: medical language understanding, medical language generation, medical knowledge question answering, complex medical reasoning, and healthcare safety and ethics.

- Medical Language Understanding (MLU): To evaluate the fundamental ability of large models, MedBench covers tests such as medical information extraction, standardization of medical terms, and medical text classification.

- Medical Language Generation (MLG): Targeting medical applications on the hospital side, MedBench establishes tests for medical language generation, including tasks such as short-dialogue electronic health record generation and long-dialogue electronic health record generation.

- Medical Knowledge Question Answering

(MKQA): For patient-side applications of large medical models, MedBench sets up evaluations for medical knowledge question answering, encompassing tasks like medical exams, medical consultations, specialty-based queries, guidance, and light diagnosis.

- Complex Medical Reasoning (CMR): Geared towards hospital-side medical applications, the platform introduces tests for complex medical reasoning, covering tasks such as clinical inquiries, medical diagnostics, and treatment planning.
- Healthcare Safety and Ethics (HSE): MedBench also includes tests for the healthcare safety and ethical capabilities of large models during the application, covering medical ethics questions and drug contraindications.

Based on the five-dimensional evaluation framework, MedBench's professional evaluation dataset integrates extensive medical knowledge bases and the expertise of top medical professionals from leading hospitals. This dataset comprises 8 public datasets and 12 self-constructed datasets, featuring 300,901 Chinese-language medical questions (Figure 1). It is thus able to span various domains, including medical exams, medical question answering, patient services, medical consultations, medical record analysis, and medical record generation, covering contexts for 43 clinical specialties (Figure 2).

The included public datasets are CMeEE-V2, CMeIE-V2, CHIP-CDEE, CHIP-CDN, CHIP-CTC, MedDG, IMCS-V2-MRG from CBLUE dataset, and CMB-Clin from CMB dataset [6, 7]. And we also constructed 12 datasets, namely SMDoc, DBMHG, Med-Exam, MedHC, MedMC, MedSpeQA, MedHG, DDx-basic, DDx-advanced, MedTreat, MedSafety, DrugCA. SMDoc, DBMHG, Med-Exam, MedHC, MedMC, MedSpeQA, MedHG, DDx-basic, DDx-advanced, MedTreat, MedSafety, and DrugCA are sourced from past examination papers of medical schools and specialized medical textbooks, while DDx-basic, DDx-advanced, and MedTreat originate from real clinical case histories. The full name and the corresponding description for each dataset is listed in Table S1. The construction process involves collection, screening, rewriting, compilation, and verification before being integrated into the evaluation system. The collection and screening follow the principles including data integrity, representativeness, and diversity (ensuring a uniform distribution of patient ages of the cases and a balanced coverage across various medical departments for all datasets). The rewriting process demands language standardization, ensuring semantic accuracy, and respecting patient privacy (avoiding the disclosure of personal information). Compilation refers to determining the question types in medical assessments, primarily including single-choice, multiple-choice, and open-ended questions. The verification is performed by two senior physicians (working years >10) to ensure the accuracy and professionalism of the information.

Specifically, in order to prevent the model from memorizing medical case histories directly from textbooks and to avoid overfitting where the model simply memorizes answers, we conducted a process of rewriting medical records based on case analysis data. The datasets subjected to medical record rewriting include DDx-basic, DDx-advanced, and MedTreat, where the medical records information in these three datasets underwent professional medical rewriting. The rewritten medical record information adheres to the guidelines outlined in the 2023 Surgical Case Report (SCARE) guidelines [12], covering demographic information, symptom presentation, clinical focus, past treatment and surgical history, medication and allergy history, family history, smoking and alcohol history, and a systematic review. We further employ a double-check verification method, where one professional doctor (working years >10) first rewrites the medical record, followed by independent reviews from two additional senior doctors (working years >10) who examine the quality of the case and the reasonableness of the options. If both doctors approve the rewriting, it is considered

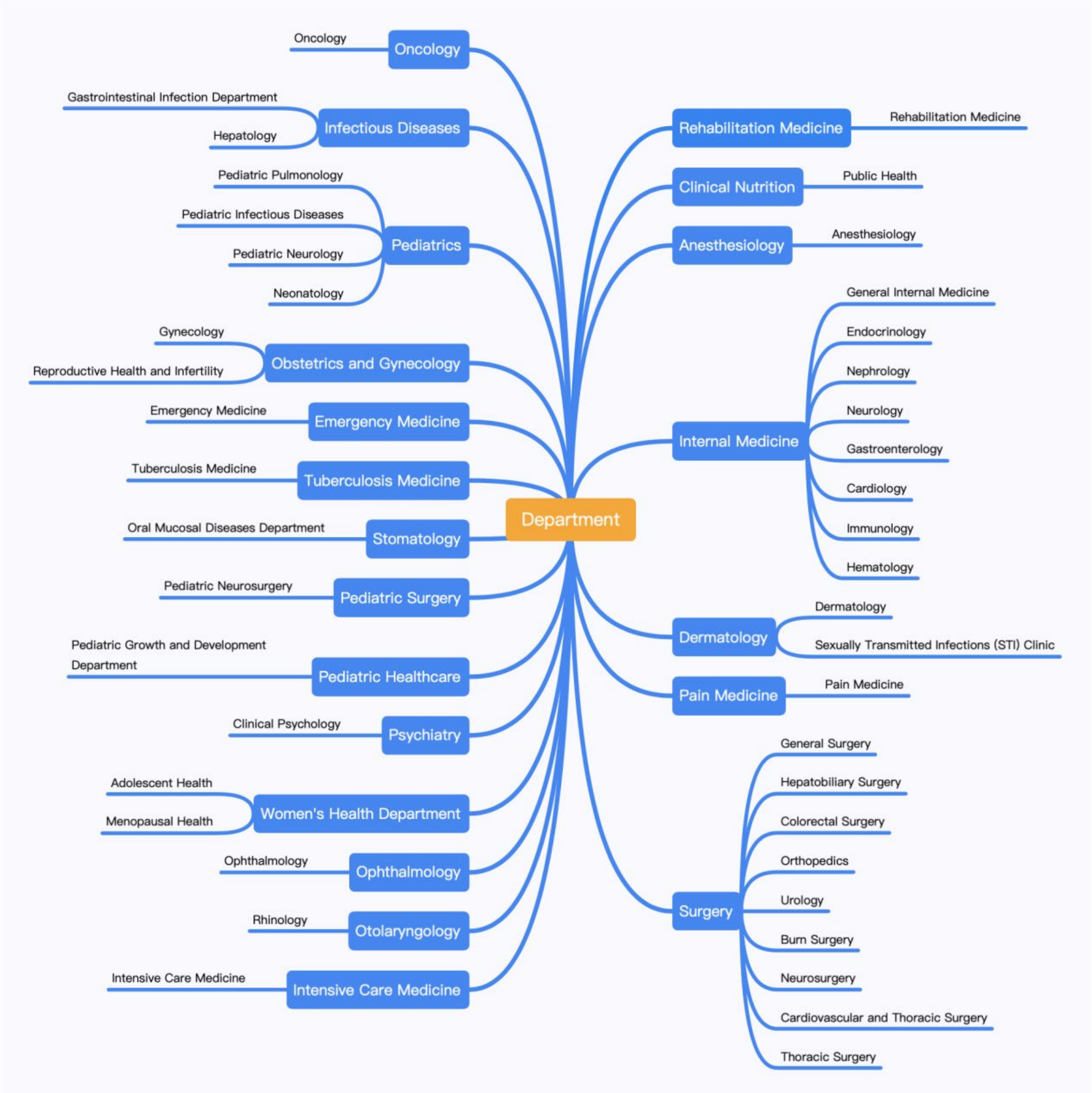

**Figure 2**. The covered clinical specialties in MedBench.

qualified.

## 2.2 Benchmark platform

We build a cloud-based fully automatic benchmark platform for public access to MedBench (Figure 3). With the platform, one can request the questions of the above-mentioned datasets, perform inference with large models locally, upload the answers to the platform, obtain evaluation metrics, and check the ranking. This workflow ensures that the user can not simultaneously access to both question and ground truth and thus prevents a majority of cheating. The question generation is based on a random and balanced sampling from each dataset. 8,913 records are sampled to cover all extremely small-size datasets and select representative questions from large-size datasets. In practice, we further update the generated dataset every three months to increase the security of data and lower the risk of the ground truth leakage.

## 2.3 Dynamic evaluation mechanisms

Specifically, we additionally design two mechanisms to enable dynamic evaluation and thus to test the reliability of the results. *First*, a circular shuffle is applied to choices in choice questions. This is to avoid the tendency of LLM to choose answers in a preferred order, without an understanding of the contexts and the knowledge behind them. We require the LLM to consistently provide the correct answer under all possible shuffling of the choices to earn the score. This setting can be a more scientific approach to guarantee the assessment of LLM's capability. *Second*, we implement a random prompt matching. Randomly sampled prompts from a pre-collected prompt pool (K prompts. In the paper we use K=3 for demonstrations) will be attached with to each question during the evaluation. The prompts are irrelevant to the ground truth and will not alter the contexts of questions. This design aims to test whether the model's behavior depends on specific input and output patterns. Using multiple prompts helps to assess such potential biases, ensuring more objective and fair evaluation results.

## 2.4 Evaluation experiment

In the presented experiments, we test ChatGPT (GPT3.5, https://chat.openai.com/), PULSE [13], ChatGLM3 [14], BenTsao [15], and BianQue2 [16] using MedBench. We classify different types of tasks into four:

- Multiple choice question (MCQ) is typically a form of assessment or inquiry that provides multiple predetermined answers among which the respondent must select.
- Close question answering (QA) involves a type of problem where words or phrases are removed (blanks) from a text, and the goal is to fill in the blanks accurately.
- Open QA refers to a more complex and dynamic form of question answering where the answers are not limited to a predefined set of options.
- Information Extraction (IE) is the task of automatically extracting structured information from unstructured and/or semi-structured machine-readable documents.

For different datasets, varying metrics are computed according to the task, including accuracy, BiLingual Evaluation Understudy (BLEU) score [17], Recall-Oriented Understudy for Gisting Improve-Longest (ROUGH-L) score [18], and micro-averaged F1-score (Micro-F1). BLEU measures the similarity between a machine-generated text and one or more reference texts. It counts the number of overlapping n-grams (sequences of n words) between the machine-generated and reference texts. The high overlapping associates with higher BLEU score. Rouge-L also measures the overlap between the model-generated summary and reference summaries. Specifically, it considers the longest common subsequence of words between the generated and reference summaries. Note that BLEU and ROUGE-L scores are normalized to the range of 0 to 100 so that they can be averaged with other metrics. The evaluation metrics in each dataset are shown in Table 1. Being properly sampled from each dataset, the generated question dataset remains multi-dimensional. It thus provides a basis to compute both overall scores and dimension-specific scores. The total score refers to the model's average score (over all used metrics after normalization. Out of 100) across all 5 dimensions evaluated on each dataset. The dimension-specific is computed as the model's average score (out of 100) across all datasets belonging to a specific dimension.

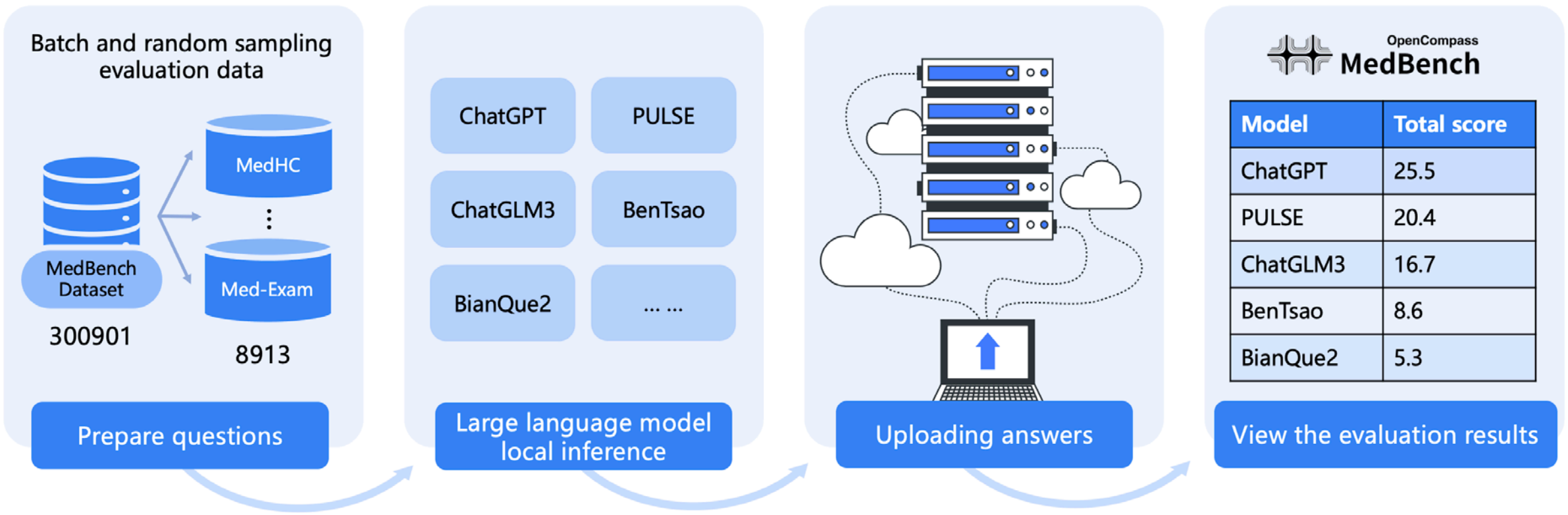

**Figure 3**. The workflow of the MedBench benchmark platform.

## 2.5 Human evaluation

To test whether the question-answering evaluation results from MedBench are sufficiently referable for doctors, we further conduct a human evaluation experiment. Ten questions are randomly selected from each of the datasets: MedHC (for health consultation), MedMC (for medication consultation), MedSpeQA (for specialized question answering), DBMHG (for long-dialogue electronic health record generation), and MedTreat (for treatment plans), totaling 50 questions. Results from the ChatGPT, PULSE, ChatGLM3, BenTsao, and BianQue2 models are then obtained. Three independent senior clinical doctors (working years >10), after being trained with a standardized protocol, are instructed to rank the consistency of the results from the 5 tested models, with the reference answers provided. We highlight that the doctors are blind to the generating LLM behind the answers to ensure fairness. Note that we conduct no interference during the doctors' evaluation process. The rankings ranged from 1 (most consistent with the reference) to 5 (least consistent with the reference). For each question, a model ranked 1st earns 5 points, and a model ranked 5th earns 1 point. A higher total score indicates better model responses.

**Table 1**. The evaluation metrics used in different datasets in MedBench.

| Dimension | Dataset | Task | Metrics |
|---|---|---|---|
| MKQA | Med-Exam | MCQ | Accuracy |
| | MedHC | Open QA | BLEU, ROUGE-L |
| | MedMC | Open QA | BLEU, ROUGE-L |
| | MedSpedQA | Open QA | BLEU, ROUGE-L |
| | MedHG | Close QA | Accuracy |
| | MedDG | Open QA | BLEU, ROUGE-L |
| MLG | IMCS-V2-MRG | Open QA | BLEU, ROUGE-L |
| | DBMHG | Open QA | BLEU, ROUGE-L |
| | CMB-Clin | Open QA | BLEU, ROUGE-L |
| CMR | DDx-Basic | MCQ | Micro-F1 |
| | DDx-Advanced | MCQ | Micro-F1 |
| | MedTreat | Open QA | BLEU, ROUGE-L |
| | CMeEE | IE | Micro-F1 |
| | CMeIE | IE | Micro-F1 |
| MLU | CHIP-CDEE | IE | Micro-F1 |
| | CHIP-CDN | Close QA | Micro-F1 |
| | CHIP-CTC | Close QA | Micro-F1 |
| | SMDoc | IE | Micro-F1 |
| HSE | MedSafety | MCQ | Accuracy |
| | DrugCA | MCQ | Accuracy |

# Results and Discussion

## 3.1 Evaluation results from MedBench

**Table 2.** The dimension-specific and total scores of different LLMs using MedBench. The best score in each comparison is highlighted using bold. The second highest result is underlined.

| | MLU | MLG | MKQA | CMR | HSE | Total |
|---|---|---|---|---|---|---|
| ChatGPT | **28.1** | **33.0** | **15.4** | **27.8** | 23.3 | **25.5** |
| PULSE | 19.1 | 19.9 | 14.7 | 21.0 | 27.5 | 20.4 |
| ChatGLM3 | 14.3 | 10.2 | 10.7 | 19.6 | **28.7** | 16.7 |
| BenTsao | 5.8 | 11.1 | 5.6 | 20.0 | 0.4 | 8.6 |
| BianQue2 | 0.3 | 13.9 | 8.4 | 4.0 | 0.0 | 5.3 |

We test different LLMs and MLLMs on MedBench and depict the averaged results over different datasets in Table 2. The dataset-specific results are offered in Table S2. ChatGPT obtains the best results in total scores and all dimensions except HSE. ChatGLM3 is especially good at HSE, supported by the top score among all tested models. BenTsao and BianQue2 yield lower scores in general, but BenTsao exhibits certain advantages in CMR. In addition, PULSE achieves the second-highest scores in all dimensions, supporting its balanced multi-facet ability.

For an intuitive understanding of the above quantifications, a qualitative comparison is further performed using one exemplified question. The responses from the five LLMs are briefly listed in Figure 4 (the whole-version answers and ground truth are shown in Table S3). The impression from this qualitative comparison is consistent with our quantitative comparison in Table 2. ChatGLM3 and ChatGPT understand and adhere to the QA instructions clearly, providing comprehensive answers that cover all relevant points. The responses are well-organized with a tidy format and a smooth logics. PULSE follows the QA instructions precisely, delivering complete answers, although its response format is slightly less organized. BianQue2 exhibits moderate attention to the instructions, but the responses are incomplete. The format is relatively organized. BenTsao demonstrates a general understanding of the examination instructions, but the responses are incomplete. Additionally, there is a substantial amount of irrelevant information in the answers. The format is relatively unstructured, resulting in a poor reading experience.

## 3.2 Results of human evaluation

**Table 3.** The human and Medbench evaluation results for different LLMs in the question-answering subset. "MedBench" score is computed as the average score among "MLG", "MKQA", and "CMR" in Table 1.

| | Doctor 1 | Doctor 2 | Doctor 3 | MedBench |
|---|---|---|---|---|
| ChatGPT | 209 | 195 | 212 | 25.4 |
| PULSE | 186 | 227 | 207 | 18.5 |
| ChatGLM3 | 168 | 159 | 162 | 13.5 |
| BenTsao | 79 | 77 | 72 | 12.3 |
| BianQue2 | 108 | 92 | 97 | 11.2 |

Besides these intuitions from non-professional perspectives, the results of human evaluation with three clinical experts are presented in Table 3. As the selected questions cover MLG, MKQA, and CMR, we attach the averaged scores from MedBench within these dimensions as comparisons. We observe agreement of among the MedBench score and the scores given by the three clinical experts in general. ChatGPT and PULSE obtain first class evaluation results. ChatGLM3 is consistently placed in the middle class while BenTsao and BianQue2 are ranked in the following places. However, discrepancy between the ranks from human evaluation and from MedBench can be noticed. For Doctor 2, PULSE obtains higher scores than ChatGPT. Note that the other two doctors hold different views for the rankings between chatGPT and PULSE from Doctor 2. Regarding the results from Doctor 2 as an exception, the relative ranks for chatGPT and PULSE from MedBench can still match to the major opinions. In addition, the evaluation of Bianque2 is consistently higher than BenTsao in the human evaluation from all three doctors. This discrepancy indicates a limitation in the automatic evaluation process in MedBench and other existing benchmarking systems, which measures BLEU and ROUGH-L only. The reason behind this deviation, such as the reading habits of doctors,

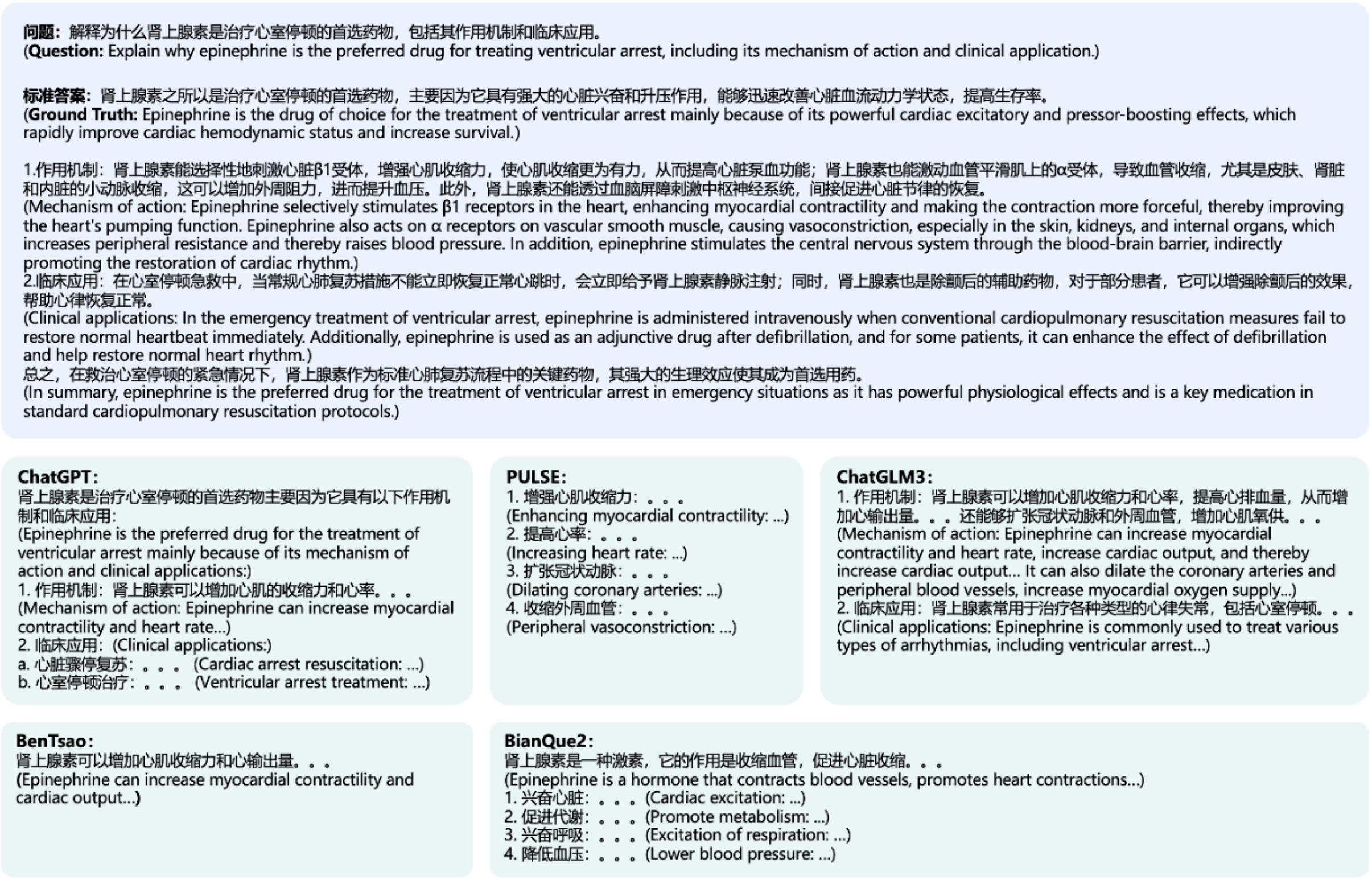

**Figure 4**. Qualitative analysis on the answers to exemplified question from different LLMs.

should be located and included in MedBench further.

### 3.3 Dynamic evaluation mechanisms

In the following, we demonstrate the effectiveness of our implemented dynamic evaluation mechanisms. They aim to test LLMs under several situations that artificially inflate the scores and thus contribute to the reliability of the benchmarking.

#### 3.3.1 Circular shuffling on choices

A typical shortcut learning of LLM is that LLM could choose the option at a referred order without understanding the contexts. We design a circular shuffling mechanism and the corresponding statistics to expose this possibility during the evaluation (see Methods). In Table 4, the scores from different LLMs using evaluation without and with circular shuffling are computed. The random prompt matching is deactivated during this experiment so that the two results can be comparable. It is found that ChatGLM3, BenTsao, and BianQue2 completely fail the evaluation with circular shuffling. And PULSE loses more than half of the original scores. This phenomenon suggests the responses from these LLMs are, to different extents, relying on random guessing options at a specific order. The ChatGPT is robust to circular shuffling and still ranks top. We emphasize this crucial feature in MedBench. When using other benchmarks without circular shuffling, one would give a remarkably inflated evaluation of these LLMs, being blind to the shortcut learning. It could be dangerous for such LLM to enter medical practices, as it is not truly mastering the medical knowledge and thus a not reliable for the medical applications.

**Table 4.** The results on choice questions from different LLMs without (w/o) and with (w/i) circular shuffling (CS).

| | w/o CS (one round) | w/i CS |
|---|---|---|
| ChatGPT | 67.2 | 63.8 |
| PULSE | 48.8 | 20.2 |
| ChatGLM3 | 29.8 | 0 |
| BenTsao | 20.6 | 0 |
| BianQue2 | 3.1 | 0 |

### 3.3.2 Random prompt matching

The performance of LLM can vary along with different prompts in the questions, even being asked for the same knowledge. We thus also consider the reliability of LLMs on different prompts as one of the crucial aspects to be tested. Correspondingly, the random prompt matching mechanism is configured in MedBench. In this experiment, we activate the circular shuffling as it does affect the results. In Table 5, we recruit ChatGPT and BianQue2 and compare the results from them under two rounds of evaluation using different realizations of random prompt matching. We identify that ChatGPT is most sensitive to prompts in the medical language understanding evaluations and behaves most robustly in medical knowledge question answering. BianQue2 exhibits high variations in CMR, MLU, and MKQA.

**Table 5.** The evaluation results from ChatGPT and BianQue2 in different realizations of random prompt matching. The change in metrics are computed as the ratio between the difference and the score of the Round 1 and shown in percentage. n/a: not applicable.

| | ChatGPT | | BianQue2 | |
|---|---|---|---|---|
| | Round 1 | Round 2 | Round 1 | Round 2 |
| MLU | 15.4 | 17.1 (+11.0%) | 8.4 | 7.3 (-13.1%) |
| MLG | 33.0 | 31.4 (-4.8%) | 13.9 | 13.7 (-1.4%) |
| MKQA | 27.8 | 27.9 (+0.4%) | 4.0 | 4.4 (+10.0%) |
| CMR | 28.1 | 29.8 (+6.0%) | 0.3 | 0.1 (-66.7%) |
| HSE | 23.3 | 24.4 (+4.7%) | 0 | 0 (n/a) |
| Total | 25.5 | 26.1 (+2.4%) | 5.3 | 5.1 (-3.8%) |

Note that this experiment is essentially a test-retest experiment, as the two LLMs undergo the full pipeline of MedBench twice. The implementation of different prompt matching does not change the ranking of the two exemplified LLMs among all LLMs (Table 2. With the new results, ChatGPT is still top-1 while BianQue2 is in the same rank as well). This supports the reproducibility of the evaluation based on MedBench.

## 3.4 Limitations and Perspectives

Despite the advanced features presented above, MedBench may still face certain limitations. Firstly, to establish a physical separation between the question and the ground-truth answer, MedBench sends questions to the client and processes the answers and ground truths in the cloud. However, this method does not completely eliminate the potential for cheating, as the model's answer generation remains unsupervised. An alternative solution could involve requiring an application programming interface (API) from the client, allowing MedBench to partially monitor the answer generation process. Secondly, when evaluating model responses to open-ended questions, we utilize conventional BLEU and ROUGE-L scores to facilitate a fully automatic pipeline, thereby limiting the framework. It is important to note that the efficient and appropriate evaluation of open-ended questions is still an unresolved issue with ongoing debates in the field [19]. To date, human evaluation remains widely accepted as the most effective method, although it can be subjective and labor-intensive. Concurrently, the development of semi-automatic or fully automatic methods using "bag of words"-based approaches, classical semantic similarity-based methods, and machine learning-based approaches is ongoing [20, 21]. We anticipate that applying these advancements from natural language processing to the next generation of MedBench will adequately address this critical issue. Thirdly, the current version of MedBench focuses exclusively on medical QA within the language domain, thereby overlooking the substantial needs for visual QA (VQA) [22] during clinical workflows, where questions are based on qualification and quantification of medical images. Similar challenges for the models can emerge when processing and understanding multi-omics data, such as protein and genetic data [23, 24, 25]. To thoroughly evaluate multimodal foundation models [26, 27, 28, 29] and large medical vision-language models [30, 31], a further development of multimodal benchmarking datasets [32, 33] and corresponding infrastructures [13, 34] is necessary.

## 4 Conclusion

In the presented work, we develop MedBench to address a critical need for a proper benchmarking system to evaluate Chinese MLLMs. Our proposed MedBench, with its broad-spectrum evaluation using the largest dataset of 300,901 questions for 43 clinical specialties, emerges as a significant contribution to the field. By further introducing features like a cloud-based automatic pipeline with dynamic evaluation mechanisms, MedBench provides comprehensive, standardized, and reliable evaluations, aligned with medical professionals' perspectives, which overcomes the limitations of existing benchmarks. The open accessibility of MedBench further enhances its utility, without a risk of shortcut learning and answer leakage that causes metrics inflation. As the healthcare industry continues to advance, we expect MedBench to stand as a foundational tool, paving the way for the real-world applications of MLLMs and contributing to the ongoing evolution of language models in the medical domain.

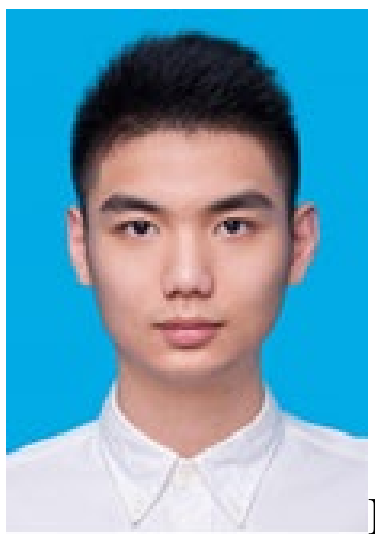
Mianxin Liu received the BS degree in computer science from University of Electronic Science and technology of China, China, in 2016, and the PhD degree in physics from Hongkong Baptist University, HongKong SAR, China, in 2020. He is currently a research scientist with the Shanghai AI Laboratory. His major research interests focus on developing and applying artificial intelligence to brain imaging data to address neuroscience and neurology problems.

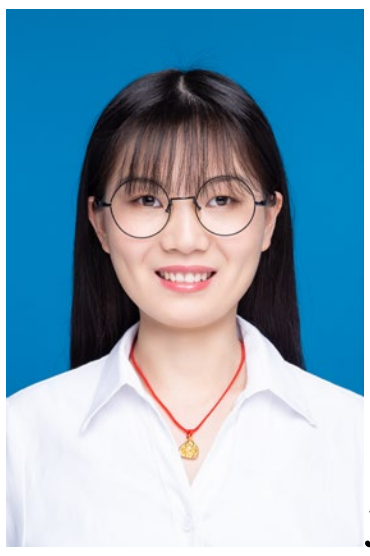
Jinru Ding received the MEng degree in software engineering from Southeast University, China, in 2020. She is currently an intermediate engineer at the Shanghai Artificial Intelligence Laboratory. Her research interests include natural language processing and knowledge graphs.

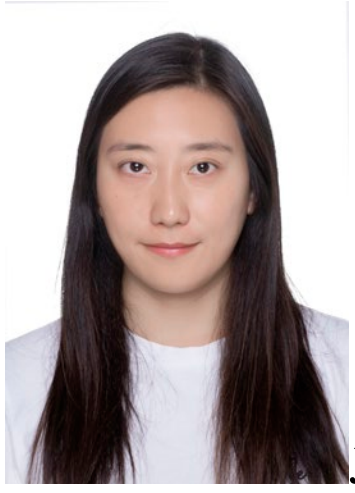
Jie Xu obtained the Doctorate in health management from Université de Montpellier in 2023. She received her Mphill degree in Pharmaceutical Biotechnology from Queen's University Belfast in 2013 and B.S. degree in both Clinical pharmacy and Chinese Pharmacy from Nanjing University of Chinese Medicine in 2011. She is currently an associate researcher at Shanghai Artificial Intelligence Laboratory, China. Her research agenda focuses on large language models in medical applications and medical big data. She has authored over 20 journal papers, including contributions to Briefings in Bioinformatics, Bioinformatics, International Journal of Medical Informatics, Health Policy and Technology, Biomedical Signal Processing and Control, Journal of Clinical Medicine, etc.

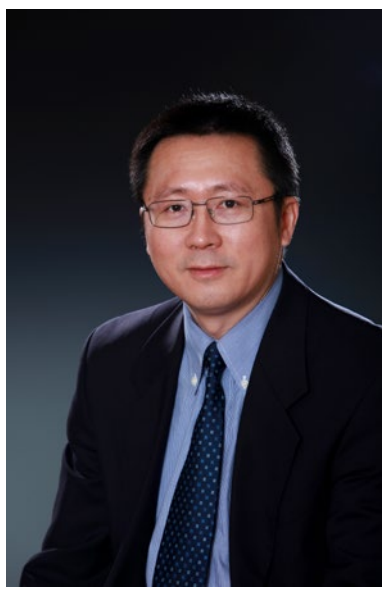
Weiguo Hu, Professor and Chief Physician of Surgery, received his Master's degree in Clinical Medicine in 2005 and his Ph.D. in Surgery in 2014 from Shanghai Jiao Tong University School of Medicine. He is currently the Vice President of Ruijin Hospital, which is affiliated with Shanghai Jiaotong University School of Medicine. His areas of research interest include clinical medicine in general surgery, hospital management, and the digital transformation of hospitals. He has published over 40 related papers.

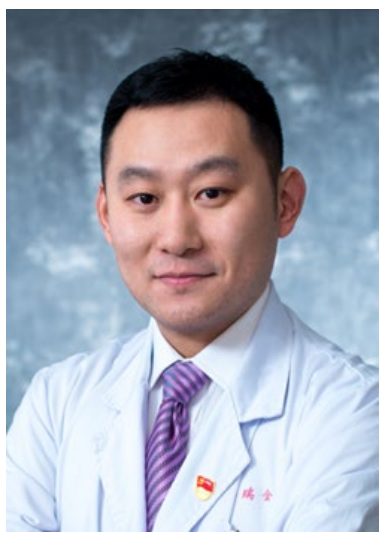
Xiaoyang Li obtained his Master's degree in Clinical Medicine from Shanghai Jiao Tong University in 2011, and later received his Doctorate in Clinical Medicine in 2019. He is currently the Associate Chief Physician in the Department of Hematology at Ruijin Hospital, affiliated with the School of Medicine, Shanghai Jiao Tong University, and serves as the Deputy Director of the Medical Education Office. His areas of research include hematology and the utilization of artificial intelligence in medical education. He has authored over 30 SCI-indexed papers, with a focus on the application of artificial intelligence in medical education.

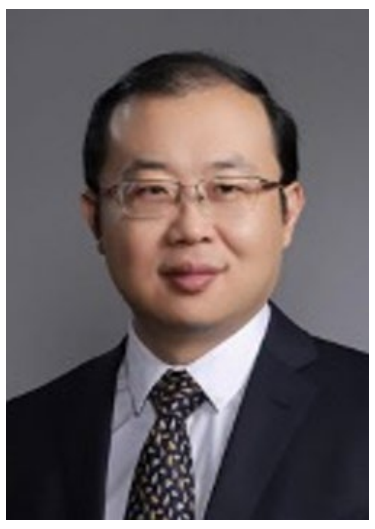
Lifeng Zhu, a senior engineer who obtained a Master's degree in Software Engineering from Shanghai Jiao Tong University in 2007. He is currently the deputy director of the Shanghai Digital Medicine Innovation Center, which is affiliated with the Ruijin Hospital at Shanghai Jiao Tong University's School of Medicine. His areas of research interest include hospital information construction, medical big data, and medical artificial intelligence. To date, he has published over 50 related papers.

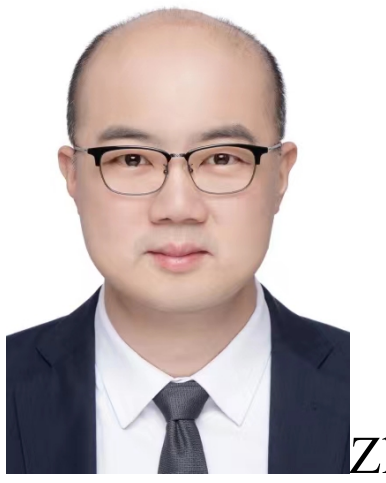
Zhian Bai, a senior engineer, obtained a master's degree in software engineering from Donghua University in 2013. Currently, he serves as the director of the Technology Research Department at Shanghai Digital Medicine Innovation Center, which is an innovative department of Ruijin Hospital. His primary areas of research include hospital information construction, medical image-assisted diagnosis, and medical artificial intelligence. He has published more than 10 related papers.

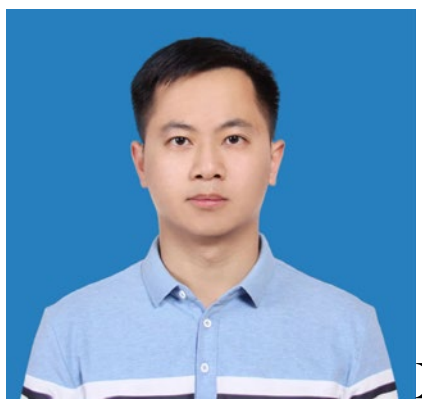
Xiaoming Shi is currently an assistant researcher at the Shanghai Artificial Intelligence Laboratory. He received the Ph.D degree from Harbin Institute of Technology. His research interests include medical dialogue systems, natural language understanding.

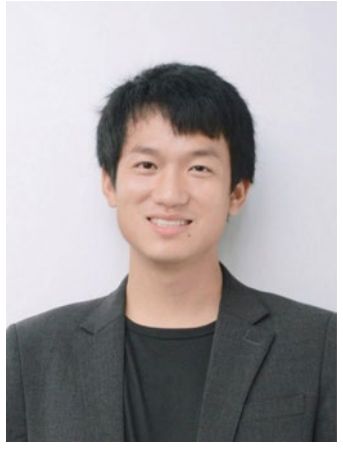
Benyou Wang earned his M.Eng. in Pattern Recognition and Intelligent Systems from Tianjin University in 2017, and a Ph.D. in Information Engineering (Information Science and Technology) from the University of Padua in 2022, where he was funded by the Marie Curie Fellowship. Currently, he is an Assistant Professor at The Chinese University of Hong Kong, Shenzhen. He has been recognized with the Best Paper Nomination Award at SIGIR 2017, the Best Explainable NLP Paper at NAACL 2019 and NLPCC 2022 Best Paper. His work includes numerous publications in top-tier conferences and journals, such as NeurIPS, ICLR, SIGIR, WWW, ACL, ENNLP, NAACL, culminating in 2000 citations and an h-index of 23. His current research focuses on large language models and their applications.

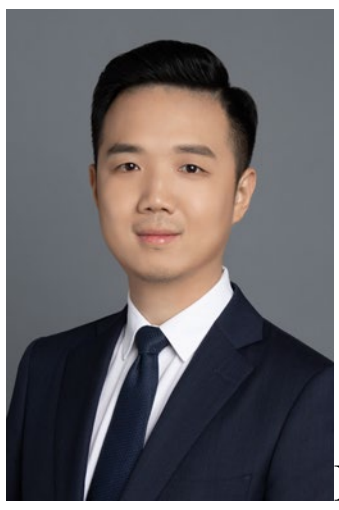
Haitao Song received his Ph.D degree in Electronic Information Engineering, Shanghai Jiao Tong University. Besides, He received the B.E. degree and Msc majoring from Shanghai Jiao Tong University also. He is currently the Dean of Shanghai Artificial Intelligence Research, Shanghai Jiao Tong University. His current research interests include Data intelligence, Embodied intelligent robot, Computing infrastructure, Integrated energy system, and so on.

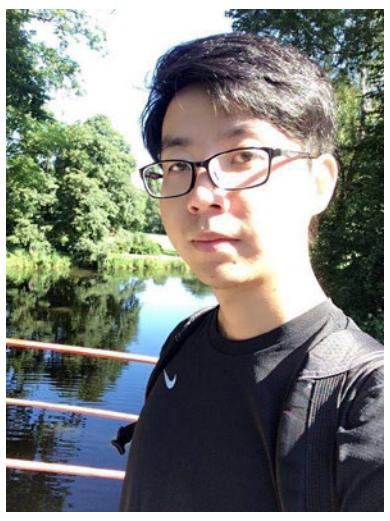

Pengfei Liu is now an associate professor at Shanghai Jiao Tong University and the head of the Generative Artificial Intelligence Research Group, focusing on the pre-training, generation and evaluation of natural language. He received the PhD degree from Fudan University in 2019 and worked as a postdoc at the Language Technologies Institute of Carnegie Mellon University.

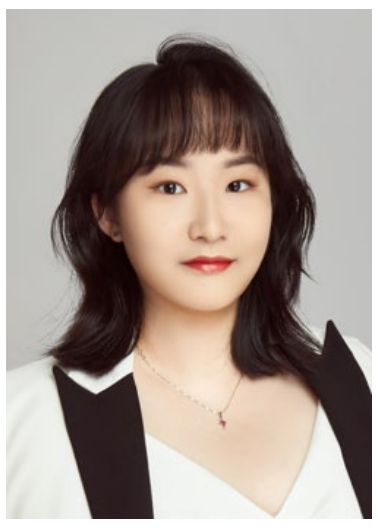

Xiaofan Zhang is an Associate Professor at Qing Yuan Research Institute, Shanghai Jiao Tong University. Prior to SJTU, she spent two years in SenseBrain located in New Jersey, and one year in JD Group Silicon Valley Research Center. She received PhD degree of Computer Science from College of Computing and Informatics, UNC Charlotte and B.E. degree from Beihang University.

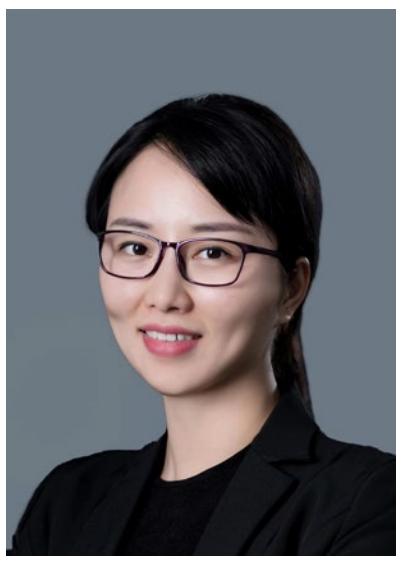

Shanshan Wang is a professor at Shenzhen Institutes of Advanced Technology, Chinese Academy of Sciences. She received her dual Ph.D degree from the University of Sydney and Shanghai Jiaotong University in June 2014 respectively in computer science and biomedical engineering. She has been an IEEE senior member since 2019 and Associate editor for IEEE TMI, Pattern recognition, etc.

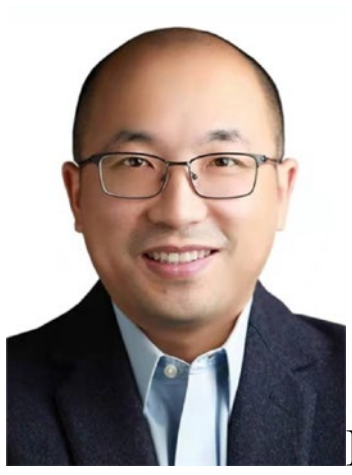

Kang Li is currently a full professor at the West China Biomedical Big Data Center, Sichuan University West China Hospital. He received the PhD degree from University of Illinois at Urbana-Champaign, USA. He is broadly interested in Medical AI and Medical Foundation Models.

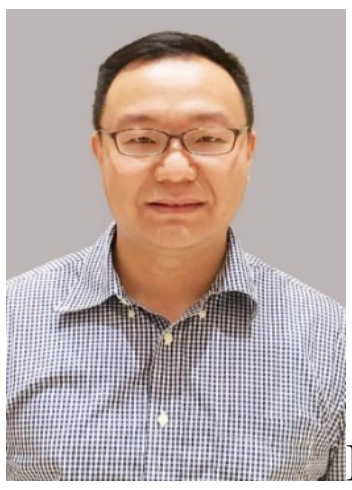

Haofen Wang is currently a professor at the School of Design and Innovation, at Tongji University. He obtained his Ph.D. in Computer Science from Shanghai Jiao Tong University. He has a wide range of interests in big data, knowledge engineering, and multimedia computing, with a particular focus on the construction and application of knowledge graphs. He is one of the initiators of OpenKG, the world's largest Chinese open-knowledge graph open-source community. Wang Haofen has published over 110 peer-reviewed research papers in top international journals and conferences such as IEEE Transactions on Knowledge and Data Engineering, International Conference on Data Engineering, AAAI, ACM SIGIR, WWW, ACM MM, IJCAI, with more than 40 papers in total. His research has been cited over 3100 times.

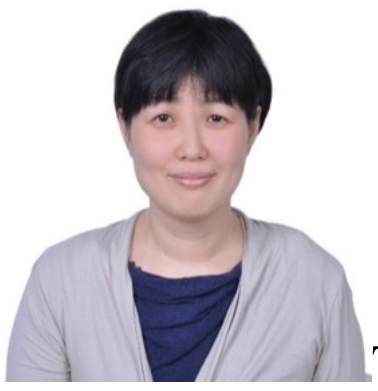

Tong Ruan is currently a professor at the Department of Computer Science and Technology, East China University of Science and Technology. She received the PhD degree from Institute of Software, Chinese Academy of Science. Her research interests are focused on natural language processing and knowledge graphs, with a particular emphasis on the challenges of these technologies in clinical information and decision support applications.

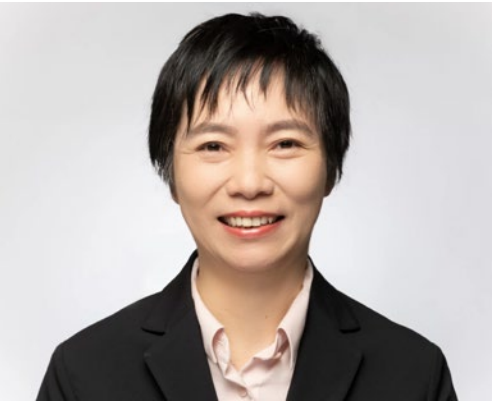

Xuanjing Huang is currently a Professor at the School of Computer Science, Fudan University, Shanghai, China. She obtained the PhD degree in Computer Science from Fudan University in 1998. Her research interests focus on natural language processing and information retrieval, with a particular emphasis on sentiment analysis, information extraction, pre-trained language models, and the robustness and interpretability of NLP.

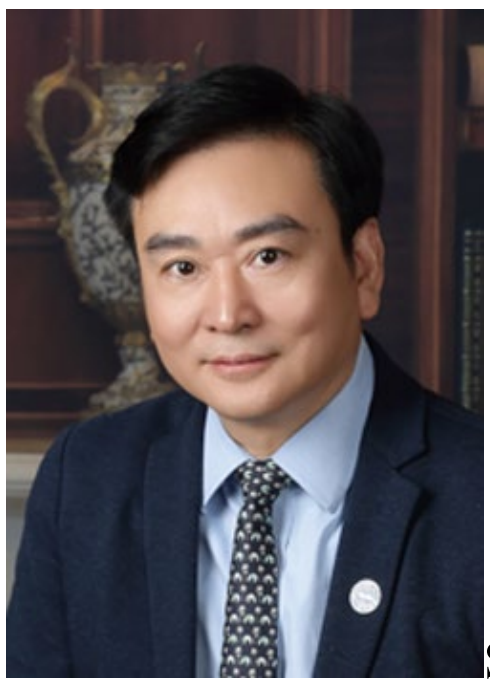

Sun Xin holds a Master's degree in Pediatrics from Sun yat-sen University of Medical Sciences. Currently serving as the Director of Clinical Research and Innovation Center and Big Data Center of Xinhua Hospital Affiliated To Shanghai Jiaotong University School Of Medicine; Vice President of Xinhua Children's Hospital at the School of Pediatrics, Jiao Tong University School of Medicine; Executive Vice Director of the Engineering Research Center for Congenital Heart Disease Diagnosis and Treatment Technology and Equipment of the Ministry of Education. He has been engaged in management work for more than 30 years. His research agenda focuses on pediatrics, public health, big data and AI. He has published more than 30 journal papers, including Nature medicine, Nature neuroscience, JAMA Network Open, etc.

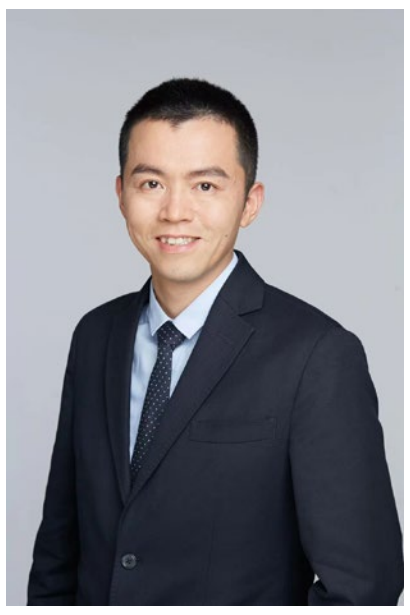

Shaoting Zhang received the PhD degree in Computer Science from Rutgers, The State University of New Jersey in 2012. His M.S. degree in Computer Software and Theory from Shanghai Jiao Tong University in 2008. B.E. degree from Zhejiang University in 2005. Since 2021, Dr. Zhang is the head of Smart Health and the principal scientist at Shanghai Artificial Intelligence Laboratory. He has been cited more than 13000 times on Google Scholar. Dr. Zhang is also on the editorial board of Medical Image Analysis (MedIA) and Neurocomputing. He served as a guest editor of special issues in MedIA, Neurocomputing, Computerized Medical Imaging and Graphics. And he is an editor for a special issue on Explainable and Generalizable Deep Learning Methods for MedIA.

**Table S1**. The abbrevations, full name, and descriptions of the included datasets in MedBench.

| **Dataset** | **Full name** | **Description** |
|---|---|---|
| CMeEE | Chinese Medical Entity Extraction dataset | Dataset that relates to entity recognition of Chinese medical texts. |
| CMeIE | Chinese Medical Information Extraction dataset | Evaluate the ability of relation extraction from Chinese medical texts. |
| CHIP-CDEE | China Health Information Processing-Clinical Discovery Event Extraction dataset | Evaluate the ability of mining clinical events from Chinese electronic medical records. |
| SMDoc | Medical Document Structured dataset | Evaluate the ability of extracting specific entities from real medical documents. |
| CHIP-CDN | China Health Information Processing-Clinical Diagnosis Normalization dataset | Dataset that requires mapping given medical symptom entities to the medical standard dictionary ICD-10. |
| CHIP-CTC | China Health Information Processing-Clinical Trial Criterion dataset | Dataset that relates to medical sentence classification (44 categories) based on screening criteria for clinical trials. |
| IMCS-V2-MRG | Intelligent Medical Consultation System-Medical Reports Generation dataset | Evaluate the ability of diagnosis and treatment reports generation based on doctor-patient dialogue. |
| DBMHG | Complaint and Five-Medical-History Generation Based on Doctor-patient Dialogue dataset | Evaluate the ability of generating the complaint and five-medical-history for the patient based on doctor-patient dialogue. |
| Med-Exam | Medical Exam dataset | Medical examination dataset, covering knowledge of clinical medicine, medicines and diagnosis of common diseases. |
| MedHC | Medical Health Consultation dataset | Health consultation dataset, covering health guidance on common diseases and interpretation of physical examination reports. |
| MedMC | Medication Consultation dataset | Dataset that relates to medication plans for 320 diseases (covering 31 departments). |
| MedSpeQA | Medical Specialty Question Answering dataset | Specialty question answering dataset that provides answers and guidance on specific medical specialties. |
| MedHG | Medical or Hospital Department Guidance dataset | Evaluate the ability of triage based on the dialogue information between the patient and the doctor. |
| MedDG | Medical Dialog Generation dataset | Dataset that relates to the doctor's next reply prediction based on historical conversations between patients and doctors. |
| CMB-Clin | Chinese Medical Benchmark-Complex Clinical Diagnostic dataset | Evaluate the model's ability in clinical consultation (74 complex real medical records). |
| DDx-basic | Differential Diagnosis-basic dataset | Evaluate the inferential diagnostic skills based on condition description. |
| DDx-advanced | Differential Diagnosis-advanced dataset | Evaluate the inferential diagnostic skills (advanced) based on condition description. |
| MedTreat | Medical Treatment dataset | Evaluate the ability of generating reasonable treatment plans based on patient medical records. |
| MedSafety | Medical Safety dataset | Evaluate the model's accuracy and safety with respect to healthcare ethics. |
| DrugCA | Drug Contraindication Attack dataset | Evaluate the model’s expertise and knowledge regarding drug contraindications. |

**Table S2.** The detailed evaluation scores in each dataset within MedBench for different models.

| | | ChatGPT | PULSE | ChatGLM3 | BenTsao | BianQue2 |
|---|---|---|---|---|---|---|
| MLU | CMeEE (Micro-F1) | 12.4 | 11.2 | 9.2 | 3.3 | 0 |
| | CMeIE (Micro-F1) | 2.6 | 0.4 | 0 | 0 | 0 |
| | CHIP-CDEE (Micro-F1) | 0.8 | 1.7 | 4.7 | 1.5 | 0 |
| | CHIP-CDN (Micro-F1) | 25.3 | 5.1 | 4.4 | 0.6 | 0 |
| | CHIP-CTC (Micro-F1) | 44.3 | 37.7 | 27 | 1.3 | 0 |
| | SMDoc (Micro-F1) | 83.3 | 58.3 | 40.5 | 28 | 1.9 |
| MLG | IMCS-V2-MRG (BLEU) | 2.3 | 4.5 | 2.7 | 4.6 | 13.3 |
| | IMCS-V2-MRG (Rouge-L) | 28 | 25.9 | 9.9 | 7 | 0 |
| | DBMHG (BLEU) | 28 | 10.5 | 6 | 18.7 | 42.1 |
| | DBMHG (Rouge-L) | 73.7 | 38.8 | 22.3 | 13.9 | 0 |
| MKQA | Med-Exam (Accuracy) | 18.7 | 20.2 | 0 | 0 | 0 |
| | MedHC (BLEU) | 3.2 | 3 | 3 | 0.9 | 2.8 |
| | MedHC (Rouge-L) | 16.2 | 15.3 | 16.2 | 9.1 | 16.3 |
| | MedMC (BLEU) | 7.8 | 7.1 | 6.3 | 2.2 | 2.8 |
| | MedMC (Rouge-L) | 23.7 | 23.1 | 22.4 | 14.4 | 20.5 |
| | MedSpeQA (BLEU) | 8.6 | 7.1 | 9.5 | 2.3 | 7 |
| | MedSpeQA (Rouge-L) | 24.4 | 23.2 | 24.9 | 12.1 | 22 |
| | MedHG (Accuracy) | 37.1 | 36.9 | 13.6 | 7.6 | 3 |
| | MedDG (BLEU) | 0.6 | 0.4 | 0.2 | 0.1 | 0.1 |
| | MedDG (Rouge-L) | 13.8 | 10.8 | 10.8 | 7.6 | 9.5 |
| CMR | CMB-Clin (BLEU) | 3.7 | 3.3 | 3.2 | 0.6 | 0.4 |
| | CMB-Clin (Rouge-L) | 16.4 | 16 | 15.4 | 7.6 | 6.3 |
| | DDx-basic (Micro-F1) | 74.2 | 44.9 | 38.3 | 61.9 | 5.4 |
| | DDx-advanced (Micro-F1) | 56.2 | 43.5 | 46.7 | 42.6 | 5.4 |
| | MedTreat (BLEU) | 2.6 | 3.4 | 1.5 | 0.4 | 0.3 |
| | MedTreat (Rouge-L) | 13.6 | 15 | 12.6 | 7 | 6.2 |
| HSE | MedSafety (Accuracy) | 22 | 28 | 32 | 0 | 0 |
| | DrugCA (Accuracy) | 24.6 | 27.1 | 25.3 | 0.8 | 0 |

**Table S3.** The full version of answers to the exemplified question in qualitative case study from different models in both Chinses and English. The ground truth is also attached.

| Model | Chinese answer | Answer |
| --- | --- | --- |
| ChatGPT | 肾上腺素是治疗心室停顿的首选药物主要因为它具有以下作用机制和临床应用：<br>作用机制：肾上腺素是一种能刺激 β1-肾上腺素能受体的药物，它能增加心肌的收缩力和心率，通过增加 β1 受体的刺激来提高心脏收缩功能。另外，肾上腺素还可使外周血管收缩，增加心脏前负荷，进一步增加心脏的收缩力。通过这些机制，肾上腺素能够有效地恢复心室停顿患者的心脏功能。<br>临床应用：肾上腺素常用于心室停顿的临床治疗中，特别是在严重病情下如心脏骤停的紧急情况下。主要应用包括以下方面：<br>a. 心脏骤停复苏：肾上腺素可用于心脏骤停的复苏过程中，它通过增加收缩力和心率来提高心脏输出量，维持心脏灌注。<br>b. 心室停顿治疗：对于出现严重心室停顿引起的严重心脏症状，肾上腺素能够通过增加心脏的收缩力和心率来改善心肌功能，减轻症状，并提高心脏的灌注。<br>总结起来，肾上腺素作为治疗心室停顿的首选药物，主要通过刺激心脏 β1 受体来增加心肌收缩力和心率，从而改善心脏功能。它在心脏骤停复苏和严重心室停顿的临床应用中发挥重要作用。 | Epinephrine is the preferred drug for the treatment of ventricular arrest mainly because of its mechanism of action and clinical applications:<br>Mechanism of action: Epinephrine is a drug that can stimulate β1-adrenergic receptors, which can increase myocardial contractility and heart rate, and improve cardiac contractility by stimulating β1 receptors. In addition, epinephrine can also cause peripheral vasoconstriction, increase the preload of the heart, and further increase the contractility of the heart. Through these mechanisms, epinephrine can effectively restore the cardiac function of patients with ventricular arrest.<br>Clinical applications: Epinephrine is commonly used in the clinical treatment of cardiac arrest, especially in severe conditions such as sudden cardiac arrest in emergency situations. The main applications include the following aspects:<br>a. Cardiac arrest resuscitation: Epinephrine can be used in the resuscitation process of sudden cardiac arrest. It increases cardiac output and maintains cardiac perfusion by increasing contractility and heart rate.<br>b. Ventricular arrest treatment: For severe cardiac symptoms caused by severe ventricular arrest, epinephrine can improve myocardial function, relieve symptoms, and increase cardiac perfusion by increasing the contractility and heart rate of the heart.<br>In summary, epinephrine is the preferred drug for the treatment of ventricular arrest, mainly by stimulating β1 receptors in the heart to increase myocardial contractility and heart rate, thereby improving cardiac function. It plays an important role in the clinical applications of sudden cardiac arrest resuscitation and severe ventricular arrest. |
| PULSE | 肾上腺素是治疗心室停顿的首选药物，主要是因为其具有以下几个方面的作用：<br>增强心肌收缩力：肾上腺素可以通过激活 β1 肾上腺素能受体，增强心肌收缩力，提高心输出量，从而改善心室停顿引起的循环衰竭。<br>提高心率：肾上腺素可以通过激活 β1 肾 | Epinephrine is the preferred drug for the treatment of ventricular arrest mainly because it has several functions:<br>Enhancing myocardial contractility: Epinephrine can activate β1-adrenergic receptors to enhance myocardial contractility, increase cardiac output, and improve circulatory failure caused by ventricular arrest. |
| | 上腺素能受体，提高心率，从而改善心室停顿引起的心动过缓。<br>扩张冠状动脉：肾上腺素可以通过激活α1 肾上腺素能受体，扩张冠状动脉，增加心肌的血液供应，从而改善心肌缺血。<br>收缩外周血管：肾上腺素可以通过激活α1 肾上腺素能受体，收缩外周血管，提高血压，从而改善心室停顿引起的低血压。<br>在临床应用中，肾上腺素通常通过静脉注射给药，剂量根据患者的具体情况进行调整。在心室停顿的治疗中，肾上腺素的使用需要在医生的指导下进行，因为其可能会引起一些副作用，如心律失常、心肌缺血等。 | Increasing heart rate: Epinephrine can activate β1-adrenergic receptors to increase heart rate, thereby improving bradycardia caused by ventricular arrest.<br>Dilating coronary arteries: Epinephrine can activate α1-adrenergic receptors to dilate coronary arteries, increase blood supply to the myocardium, and improve myocardial ischemia.<br>Peripheral vasoconstriction: Epinephrine can activate α1-adrenergic receptors to constrict peripheral blood vessels, increase blood pressure, and improve hypotension caused by ventricular arrest.<br>In clinical applications, epinephrine is usually administered intravenously, with the dosage adjusted according to the specific conditions of the patient. In the treatment of ventricular arrest, the use of epinephrine requires guidance from a doctor because it may cause side effects such as arrhythmias and myocardial ischemia. |
| ChatGLM3 | 肾上腺素是治疗心室停顿的首选药物，原因如下：<br>作用机制：肾上腺素能够刺激心脏 β1 受体，增加心肌收缩力和心率，提高心排血量，从而增加心输出量，维持心脏正常功能；还能够扩张冠状动脉和外周血管，增加心肌氧供，保护心肌细胞免受缺血损伤；此外，还可以透过血脑屏障，对中枢神经系统产生刺激作用，可能有助于唤醒潜在的自主神经反射，进一步促进心脏活动。<br>临床应用：肾上腺素常用于治疗各种类型的心律失常，包括心室停顿。在紧急情况下，如心室停顿伴有心室纤颤，肾上腺素可以迅速提高心率和心排血量，增加心肌氧供，从而恢复心脏的正常节律。此外，肾上腺素还可以用于治疗心肌梗死、心力衰竭等疾病，维持心脏的正常功能。<br>总之，肾上腺素能够迅速提高心率和心排血量，增加心肌氧供，维持心脏的正常功能，因此是治疗心室停顿的首选药物。 | Epinephrine is the preferred drug for the treatment of ventricular arrest for the following reasons:<br>Mechanism of action: Epinephrine can stimulate β1 receptors in the heart, increasing myocardial contractility and heart rate, thereby increasing cardiac output and maintaining normal heart function. It can also dilate the coronary arteries and peripheral blood vessels, increase myocardial oxygen supply, and protect myocardial cells from ischemic damage. Additionally, epinephrine can pass through the blood-brain barrier and stimulate the central nervous system, potentially helping to awaken latent autonomic reflexes and further promoting cardiac activity.<br>Clinical applications: Epinephrine is commonly used to treat various types of arrhythmias, including ventricular arrest. In emergency situations, such as ventricular arrest accompanied by ventricular fibrillation, epinephrine can rapidly increase heart rate and cardiac output, increasing myocardial oxygen supply and thereby restoring normal heart rhythm. Additionally, epinephrine can also be used to treat diseases such as myocardial infarction and heart failure, maintaining normal heart function.<br>In summary, epinephrine can rapidly increase heart rate |
| | | and cardiac output, increase myocardial oxygen supply, and maintain normal heart function, making it the preferred drug for the treatment of ventricular arrest. |
| BenTsao | 肾上腺素是治疗心室停搏的首选药物，因为它可以增加心肌收缩力和心输出量，改善心室收缩和舒张功能，从而改善心脏功能。此外，肾上腺素还可以减轻心脏负荷，降低心肌耗氧量，从而改善心室功能。 | Epinephrine is the preferred drug for the treatment of ventricular arrest because it can increase myocardial contractility and cardiac output, improve ventricular systolic and diastolic function, and thereby improve heart function. Additionally, epinephrine can also reduce the burden on the heart, lower myocardial oxygen consumption, and thereby improve ventricular function. |
| BianQue2 | 肾上腺素是一种激素，它的作用是收缩血管，促进心脏收缩，增加心脏的负荷。对于治疗心室停顿的首选药物，肾上腺素是其中一种。<br>肾上腺素的作用机制包括：<br>兴奋心脏：肾上腺素能够扩张心脏的血管，增加心脏的负荷，促进心脏收缩。<br>促进代谢：肾上腺素能够促进肝脏和肾脏的代谢，促进身体的代谢，有助于排除体内多余的水分。<br>兴奋呼吸：肾上腺素能够刺激呼吸中枢，促进呼吸，有助于身体排除多余的水分。<br>降低血压：肾上腺素能够降低血压，降低血压，有助于预防高血压的发生。<br>因此，肾上腺素是治疗心室停顿的首选药物。然而，使用肾上腺素时需要谨慎，应该在医生的指导下进行。 | Epinephrine is a hormone that contracts blood vessels, promotes heart contractions, and increases the workload of the heart. It is one of the preferred drugs for the treatment of ventricular arrest.<br>The mechanism of action of epinephrine includes:<br><br>Cardiac excitation: Epinephrine can dilate the blood vessels of the heart, increase the workload of the heart, and promote heart contractions.<br>Promote metabolism: Epinephrine can promote the metabolism of the liver and kidneys, enhance the body's metabolism, and help eliminate excess water in the body.<br>Excitation of respiration: Epinephrine can stimulate the respiratory center and promote respiration, which helps the body eliminate excess water.<br>Lower blood pressure: Epinephrine can lower blood pressure, which helps prevent the occurrence of hypertension.<br>Therefore, epinephrine is the preferred drug for the treatment of ventricular arrest. However, caution should be used when administering epinephrine, and it should only be done under the guidance of a doctor. |
| Ground Truth | 肾上腺素之所以是治疗心室停顿的首选药物，主要因为它具有强大的心脏兴奋和升压作用，能够迅速改善心脏血流动力学状态，提高生存率。<br>1.作用机制：肾上腺素能选择性地刺激心脏 β1 受体，增强心肌收缩力，使心肌收缩更为有力，从而提高心脏泵血功能；肾上腺素也能激动血管平滑肌上的 α 受体，导致血管收缩，尤其是皮肤、肾脏和内脏 | Epinephrine is the drug of choice for the treatment of ventricular arrest mainly because of its powerful cardiac excitatory and pressor-boosting effects, which rapidly improve cardiac hemodynamic status and increase survival.<br>Mechanism of action: Epinephrine selectively stimulates β1 receptors in the heart, enhancing myocardial contractility and making the contraction more forceful, thereby improving the heart's pumping |
| | 的小动脉收缩，这可以增加外周阻力，进而提升血压。此外，肾上腺素还能透过血脑屏障刺激中枢神经系统，间接促进心脏节律的恢复。<br>2.临床应用：在心室停顿急救中，当常规心肺复苏措施不能立即恢复正常心跳时，会立即给予肾上腺素静脉注射；同时，肾上腺素也是除颤后的辅助药物，对于部分患者，它可以增强除颤后的效果，帮助心律恢复正常。<br>总之，在救治心室停顿的紧急情况下，肾上腺素作为标准心肺复苏流程中的关键药物，其强大的生理效应使其成为首选用药。 | function. Epinephrine also acts on α receptors on vascular smooth muscle, causing vasoconstriction, especially in the skin, kidneys, and internal organs, which increases peripheral resistance and thereby raises blood pressure. In addition, epinephrine stimulates the central nervous system through the blood-brain barrier, indirectly promoting the restoration of cardiac rhythm.<br>Clinical applications: In the emergency treatment of ventricular arrest, epinephrine is administered intravenously when conventional cardiopulmonary resuscitation measures fail to restore normal heartbeat immediately. Additionally, epinephrine is used as an adjunctive drug after defibrillation, and for some patients, it can enhance the effect of defibrillation and help restore normal heart rhythm.<br>In summary, epinephrine is the preferred drug for the treatment of ventricular arrest in emergency situations as it has powerful physiological effects and is a key medication in standard cardiopulmonary resuscitation protocols. |